\documentclass{article}
\usepackage{spconf,amsmath,graphicx,hyperref}

\title{Making Dialogue Grounding Data Rich: A Three-Tier Data Synthesis Framework for Generalized Referring Expression Comprehension}
\name{Juexi Shao$^{1}$, Siyou Li$^{1}$, Yujian Gan$^{2*}$\thanks{$^{*}$Corresponding author},
    Christopher Madge$^{1}$, Vanja Karan$^{3}$, Massimo Poesio$^{1,4}$}
\address{$^{1}$ Queen Mary University of London, London, UK\\
        $^{2}$ Queen's University Belfast, Belfast, UK\\
        $^{3}$ University of Vienna, Vienna, Austria \\ $^{4}$ Utrecht University, Utrecht, Netherlands
      }
      
\usepackage[colorinlistoftodos]{todonotes}

\usepackage{enumitem}  
\usepackage{multirow}
\usepackage{xcolor,amssymb}
\usepackage[framemethod=tikz]{mdframed} 

\usepackage[ruled,vlined,linesnumbered,noend]{algorithm2e}
\SetKwInput{KwIn}{Input}
\SetKwInput{KwOut}{Output}

\usepackage{caption}
\begin{document}
%
\maketitle
\small\begin{abstract}
Dialogue-Based Generalized Referring Expression Comprehension (GREC) requires models to ground the expression and unlimited targets in complex visual scenes while resolving coreference across a long dialogue context. 
However, existing systems struggle under distribution shift between training and evaluation domains, a gap exacerbated by the scarcity of annotated dialogue grounding data.
We address this challenge with a three-tier data-synthesis method that balances realism and controllability to produce scalable supervision for dialogue-conditioned grounding. Fine-tuning on the synthesized data yields consistent, substantial improvements over prior approaches across standard evaluation metrics. The code and prompts are available at \url{https://github.com/jshao-cs/MDGDR}.
\end{abstract}
\small\section{Introduction}
\label{sec:intro}
Referring Expression Comprehension (REC) - the task of locating a target referred to by a natural language description ~\cite{kazemzadeh2014referitgame,yu2016modeling,plummer2015flickr30k} is key in vision-language research. 
Recent advances have pushed the state of the art beyond simple surface matching toward richer use of semantic information \textemdash
most notably, constructing compositional referring expression (RE)~\cite{chen2020cops} and incorporating external knowledge~\cite{wang2020give,chen2023advancing,madge2025mdc}. Generalized REC (GREC) ~\cite{he2023grec,madge2025mdc} -- the task of interpreting REs referring to zero or multiple objects, as in \textit{Pick up those three boxes} -- relaxes these assumptions to better reflect real-world communication, where the referent may be uncertain, occluded, or missing from the scene.

However, this research typically relies on substantial collections of real-world images with large volumes of human annotations. 
Online crowdsourcing platforms 
offer controllable, reproducible testbeds for visual grounding research,  helping to mitigate data collection costs and variability ~\cite{narayan2019collaborative,kiseleva2022interactive}. 
Yet the situation changes when models are asked to jointly perceive the scene, maintain discourse state, and interact over multiple turns: under these conditions, existing approaches begin to show their 
limits. 
~\cite{madge2025mdc} introduced the first GREC benchmark grounded in multi-turn dialogue within a Minecraft building environment, 
finding that this task poses a formidable challenge to both classical transformer-based neural networks and Large Vision-Language Models (LVLMs). 
\cite{madge2025mdc} also found that the cost of dialogue annotation is 
very high, 
resulting in an absence of 
in-domain training data and making the acquisition of large-scale training datasets a long-term goal. 
This motivates the data augmentation framework proposed in this paper, which synthesizes concise REs and rich dialogues to scale grounded comprehension and stepwise reasoning.

\begin{figure}
    \centering 
    \includegraphics[width=\columnwidth]{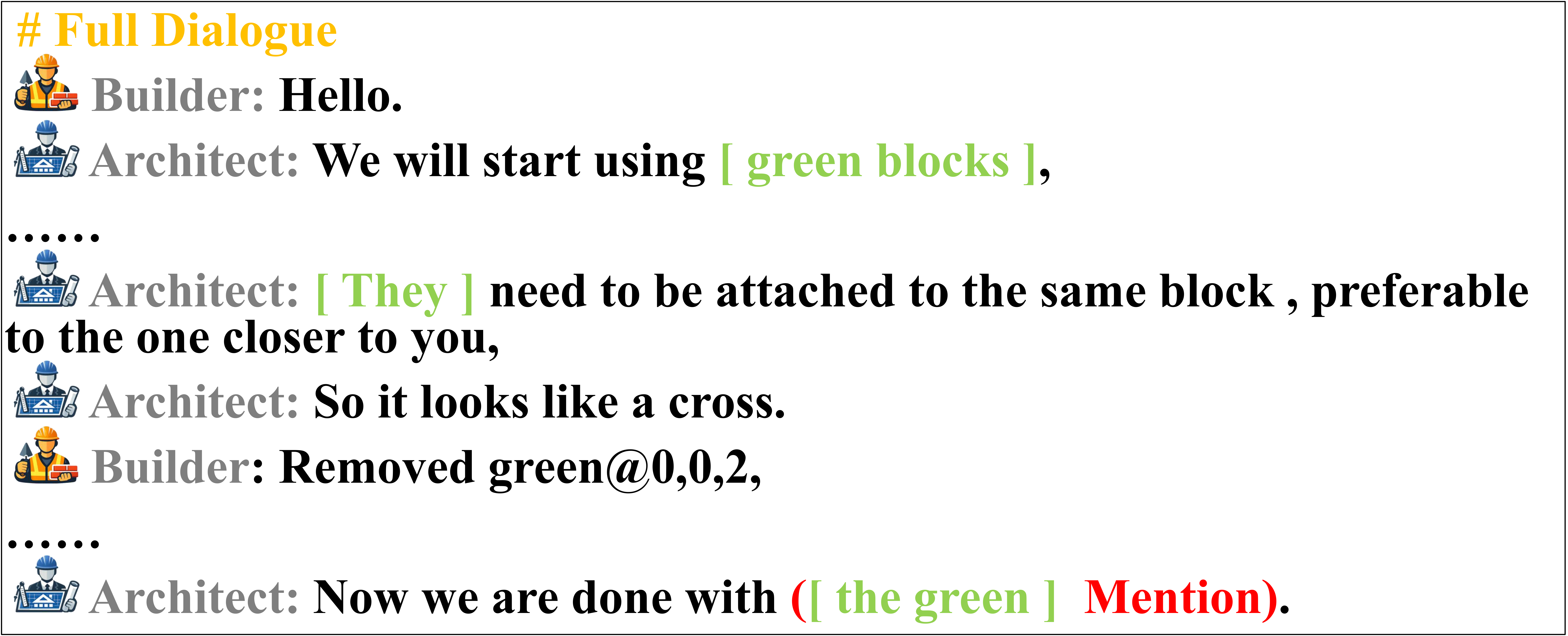} 
    \includegraphics[width=\columnwidth]{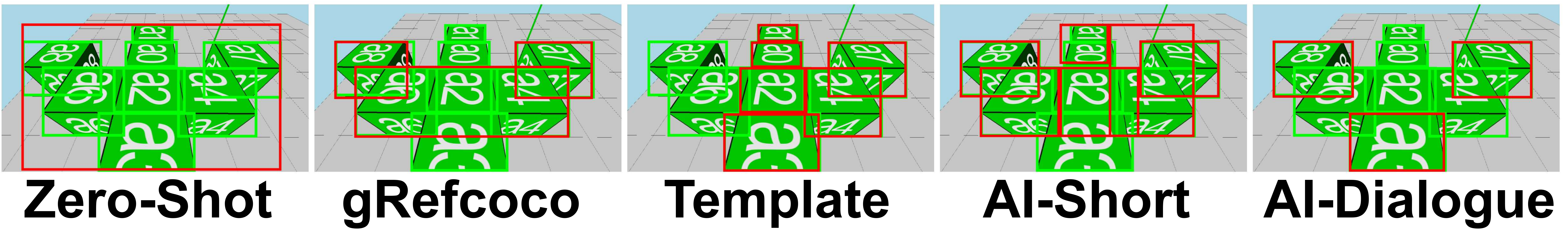} 
    \caption{Given a \textbf{Full Dialogue} and a Minecraft world \textbf{Image}, the model needs to recognize all blocks that refer to the \textbf{Mention}. The coreference chain of the full dialogue is highlighted in green, and inference results of Qwen2-VL fine-tuned on various data are shown below. All the green bounding boxes represent the ground truth, while the red bounding boxes represent the prediction.}
    \label{fig::case} 
\end{figure}

Prior work on AI-driven data generation is relevant to our setting~\cite{islam2022caesar,yumetamath,lu-etal-2024-mathgenie}. Yet translating these successes to GREC remains challenging: beyond fluent text synthesis, the task demands precise modeling of multi-object spatial relations and faithful grounding within the visual scene.

We address the scarcity of MDC-R~\cite{madge2025mdc} data by synthesizing data at three tiers of complexity to better reflect varying degrees of realism and interactional complexity. 
At the simplest tier, we generate single-turn RE using a template-based method. 
At the intermediate tier, we employ GPT-4 to produce a broader spectrum of compositional REs (still single utterance).  
At the highest tier, we synthesize full multi-turn dialogues that evolve up to the moment of reference. 
Together, these sources constitute a ladder of expressiveness\textemdash from controlled single-turn mentions to dialogue references that highlight short-term memory, clarification, and disambiguation.


\noindent\textbf{In summary, we make three contributions:}
(1) A three-tier data augmentation framework for solving the data sparsity of MDC-R ~\cite{madge2025mdc}, spanning short REs to multi-turn dialogues.
(2) Experimental demonstration that the model trained on these synthetic data achieves notable improvements, with precision increasing by \(\approx 20\%\).
(3) The finding that biases across data types influence model learning and generalization, motivating the importance of distribution-aware training.

\small\section{Related Work}
\label{sec:format}
\textbf{REC and GREC} REC has advanced rapidly recently. The early two-stage paradigm~\cite{yu2018mattnet}, couples off-the-shelf object detectors with linguistic cues to compute region–expression matching scores. The field has progressed from specialist to generalist~\cite{kamath2021mdetr}, which are pre-trained at scale to jointly learn detection, alignment, and reasoning, and, more recently, to LVLMs~\cite{wang2024qwen2}. Despite strong results on classical benchmarks, these systems often struggle in realistic settings use\textemdash e.g., partially specified scenes where external knowledge must be integrated for reasoning ~\cite{wang2020give,chen2023advancing,madge2025mdc}.

GREC extends REC by requiring models to localize an unlimited number of targets per RE. Most current approaches adopt transformer-based architectures ~\cite{he2023grec,hemanthage2024recantformer}; some LVLMs are adapted via integrating GREC data ~\cite{pramanick2024jack}. Nevertheless, performance remains far from saturated.

\noindent\textbf{The Minecraft Dialogue Corpus} MDC-R~\cite{madge2025mdc} consists of dialogues~\cite{narayan2019collaborative} between an architect, who gives instructions, and a builder, who constructs a structure in a virtual world, as illustrated in Fig.~\ref{fig::template}.

\noindent\textbf{Multimodal Data Synthesis} As the data demands for large-scale pre-training have grown, synthetic data has emerged as a prominent paradigm. A broad class of methods ~\cite {narayan2019collaborative,kiseleva2022interactive,shridhar2020alfred} relies on simulators to produce data with controllable distributions and programmatically generated annotations ~\cite{islam2022caesar,9010706,liu2019clevr}, whereas recent approaches leverage generative models to synthesize diverse textual ~\cite{yumetamath,lu-etal-2024-mathgenie} and multimodal ~\cite{parolari2025harlequin} training signals. Both lines of work aim to expand coverage and reduce manual labelling effort. Based on previous works, we propose three distinct tiers of data synthesis methods.

\small\section{Methodology}
\label{sec:methodology}
To balance data realism and controllability, we introduce a three-tier synthesis framework comprising: (i) template-based short RE synthesis, (ii) prompted short RE synthesis, and (iii) full dialogue with coreference information synthesis. 
We detail the construction procedure and explain how the components integrate to form the final training corpus. Prior to RE synthesis, we simulate screenshots across 101 scenes and compute bounding boxes for all blocks. Details are provided in Section~\ref{sec:methodology:annotation}.

\subsection{Short Referring Expressions}
\subsubsection{Template-based Short RE Generation}
\label{sec:methodology:template}
The MC virtual environment we use features an 11 x 11 grid where seven different colored blocks can be freely placed, enabling an infinite number of possibilities for constructing target structures. Selecting the appropriate block from an environmental screenshot has become a challenging task, as entity mentions must be unambiguous. A heuristic approach involves grouping all blocks by color into distinct sets, then progressively creating mention templates by constructing different attributes.
Building upon previous work ~\cite{islam2022caesar}, we develop several different compositional templates.

\begin{table}[htb]
\centering
\resizebox{\columnwidth}{!}{
\begin{tabular}{l|l}
    \hline
    Template & Example \\
    \hline
    the \textless Color \textgreater{} block(s)& the red blocks \\
    the \textless Color \textgreater{} \textless Geometric Shape \textgreater{} & the red bar \\
    the \textless position \textgreater{} of the \textless Color \textgreater{} \textless Geometric Shape \textgreater{} & the first block of the yellow column\\
    \hline
\end{tabular}
}
\caption{ Templates for the short compositional RE.
}
\label{tab::template}
\end{table}


\begin{figure}[ht]
    \centering 
    \includegraphics[width=0.55\columnwidth]{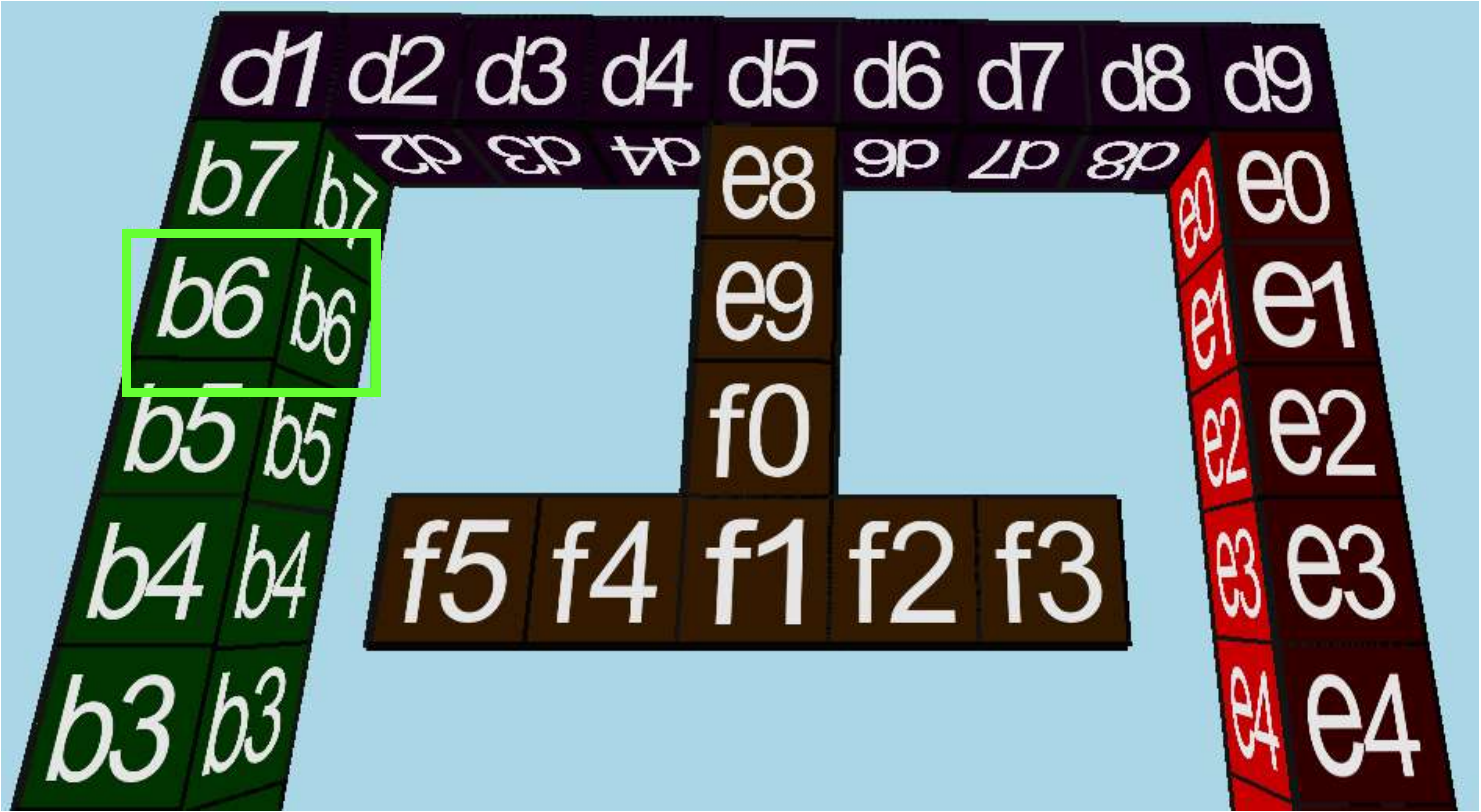} 
    \caption{The MC environment - example of the image of synthetic RE 'the second green block from the top'.}
    \label{fig::template} 
\end{figure}

Taking the second template as an example, we can create a RE for “the second green block from the top” based on the arrangement of green blocks in the Fig. \ref{fig::template}, and it's unambiguous.

\subsubsection{Prompted Synthetic RE Generation}
\label{sec:methodology:prompted}
Prior studies ~\cite{yumetamath,lu-etal-2024-mathgenie} have used Large Language Models (LLMs) to synthesize constrained training data by Chain-of-Thought (CoT). In our setup, we randomly sample a target block via a moderately sized crop and prompt GPT\textendash 4 to extract its attributes (e.g., \textit{id}, color, position, ordinal, reference, perspective). We then instantiate two controlled templates to compose a single, unambiguous RE per target, yielding compositional supervision for downstream training.

We synthesize compositional REs with GPT-4.1~\cite{openai-gpt41-api-2025} using a tightly constrained instruction. Each query provides a pair of images: (i) the full Minecraft scene and (ii) a tight crop centered on the target block. The model is required to (a) recover the block identifier \texttt{id} from the crop and (b) produce a single, well-formed RE that uniquely localizes the target within the full scene.

To ensure linguistic consistency and downstream parsability, the model must instantiate exactly one of two canonical patterns, each ending with a period:


\begin{mdframed}
(i) the center \textless color\textgreater{} block of the \textless reference\textgreater{}, \textless perspective\textgreater{}.
\end{mdframed}
\begin{mdframed}
(ii) the \textless ordinal\textgreater{} \textless color\textgreater{} block from the \textless position\textgreater{} of the \textless reference\textgreater{}, \textless perspective\textgreater{}.
\end{mdframed}

The semantic slots are constrained as follows:
\begin{itemize}[]
\item \textbf{\texttt{reference}}: a noun phrase denoting the larger structure containing the target (e.g., ``arch'').
\item \textbf{\texttt{position}}: one or more words indicating the position of the target within
\texttt{reference}.
\item \textbf{\texttt{ordinal}}: a ordinal number indexing target order within \texttt{reference} (such as, ``first'' and ``second'').
\item \textbf{\texttt{color}}: the pixel color of the target.
\item \textbf{\texttt{perspective}}: one supplemental modifier  (e.g., ``close to you'') defined relative to the builder's viewpoint in the full scene.
\end{itemize}

Additionally, GPT-4 must output an explicit ID to verify target identification. The instruction requests CoT reasoning for reliability, while constraining the output to the final JSON only; this closed-form grammar and typed schema reduce variance, enabling strict validation. The short synthetic REs are then extracted from the JSON.

\subsection{Synthesizing Dialogues with Coreference}
\label{sec:methodology:dialogue}

We target the scarcity of GREC datasets featuring \emph{multi-turn, co-referential} dialogues. Although LLMs can be prompted to synthesize dialogues, recent work ~\cite{gan2024assessing} indicates that off-the-shelf LLMs exhibit limited coreference tracking. We therefore fine-tune Qwen2-VL~\cite{wang2024qwen2} on external coreference-aware dialogue VisPro corpora~\cite{yu2019you} to learn coherent dialogue generation with explicit coreference chains in Minecraft scenes. The outputs contain (i) coreference-consistent dialogues and (ii) structured REs, providing complementary supervision for model training.

\begin{figure}[ht]
    \centering 
    \includegraphics[width=\columnwidth]{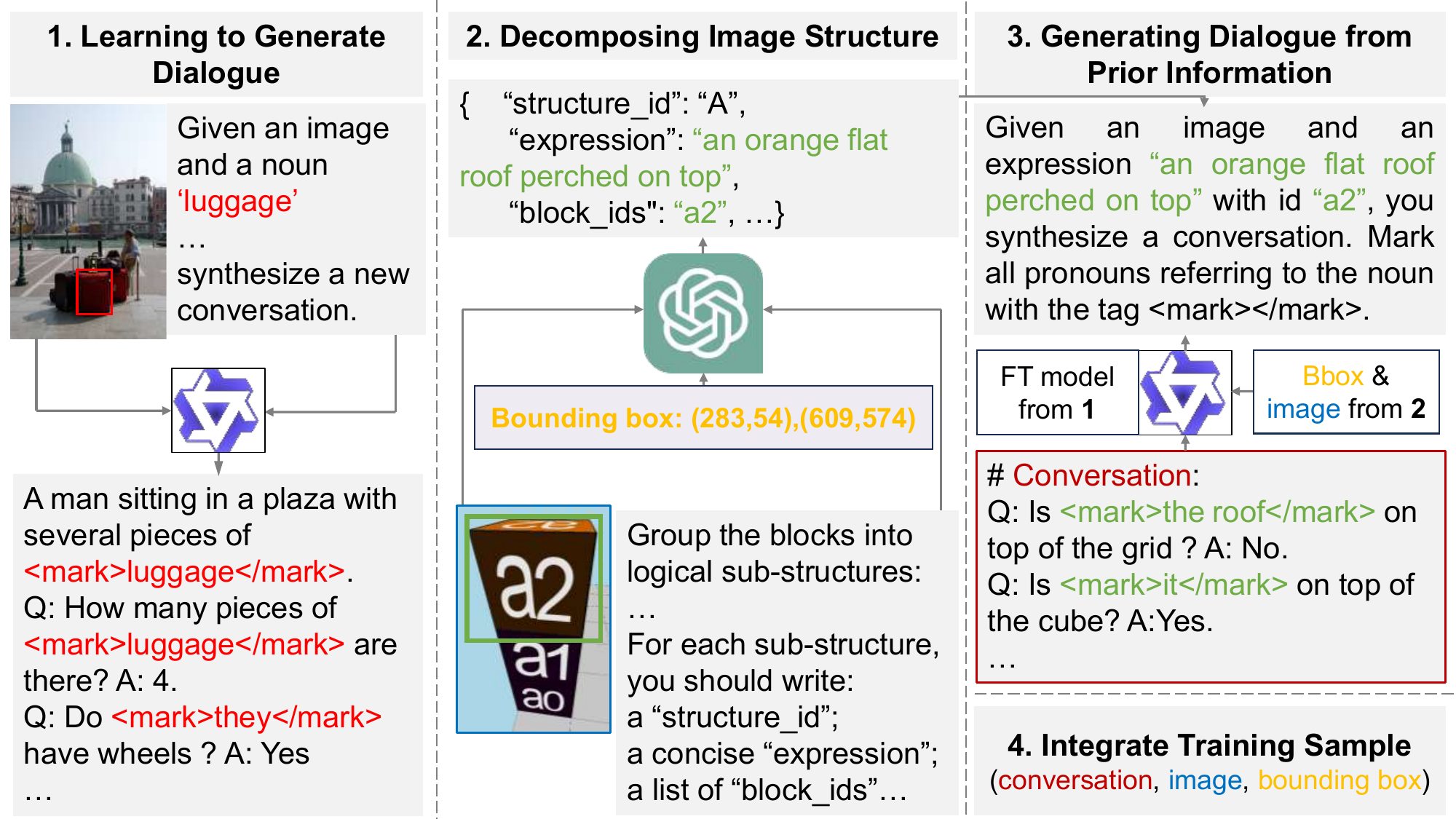} 
    \caption{Method of generating multi-turn dialogue containing coreference chain.}
    \label{brainstorm} 
\end{figure}
In particular, our method comprises four stages as shown in Fig.~\ref{brainstorm}.

(1) \textbf{Learning to Generate Dialogue} We fine-tune Qwen2-VL $\mathcal{M}$ (with parameter $\theta$) with a LoRA Adaptor (consisting of low rank matrices $\mathbf{A}\in\mathbb{R}^{d\times r},\mathbf{B}\in\mathbb{R}^{r\times d}$) on a visual coreference corpus to learn the mapping from (image $\mathcal{I}_\mathrm{ext}$, bounding box on the target $\mathcal{B}_\mathrm{ext}$ and RE $\mathcal{E}_\mathrm{ext}$) to a coherent multi-turn, coreference-consistent dialogue $\mathcal{D}_\mathrm{ext}$, and preserve a model $\mathcal{M}^{\mathrm{'}}$ with new parameter \renewcommand\theequation{\alph{equation}}
\begin{equation}
\theta^{\mathrm{ft}} \;:=\theta + \Delta\theta , \: \mathrm{where} \: \Delta\theta =\mathbf{A}\mathbf{B}.
\end{equation}
(2) \textbf{Decompose Image Structure} For each simulated Minecraft scene image $\mathcal{I}_\mathrm{gt}$, we decompose the scene into distinct substructures using GPT-4.1 $\mathcal{M}^{\mathrm{GPT}}$, produce REs $\mathcal{E}$ with a list of unique block IDs $\mathcal{U}_\mathrm{gt}$ as the identifier, and align bounding boxes for the block IDs $\varphi:\;\mathcal{B}_\mathrm{gt} \longrightarrow \mathcal{U}_\mathrm{gt}$ in every substructure programmatically:
\renewcommand\theequation{\alph{equation}}
\begin{equation}
    (\mathcal{U}_\mathrm{gt},\mathcal{E}):=\; \mathcal{M}^{\mathrm{GPT}}({\mathcal{I}_\mathrm{gt})}.
\end{equation}
(3) \textbf{Generating Dialogue from Prior Information} We condition the fine-tuned Qwen2-VL $\mathcal{M}^{\mathrm{'}}$ on the simulated image $\mathcal{I}_\mathrm{gt}$ and the derived REs $\mathcal{E}$ from (2) to generate RE-based dialogues $\mathcal{D}_{\mathcal{E}}$ with coreference information:
\renewcommand\theequation{\alph{equation}}
\begin{equation}
    \mathcal{D}_{\mathcal{E}}:=\; \mathcal{M}^{\mathrm{'}}(\theta^{\mathrm{ft}};\,\mathcal{E},\mathcal{I}_\mathrm{gt}).
\end{equation}
(4) \textbf{Integrate Training Sample} We integrate the dialogue, image and aligned bounding boxes into a single training sample $( \mathcal{D}_{\mathcal{E}},\mathcal{I}_\mathrm{gt},\mathcal{B}_\mathrm{gt})$ for GREC training.

\small\section{Experiment}
\label{sec:experiment}

\subsection{Dataset}
\label{sec:experiment:dataset}
We adopted the MDC\textendash R benchmark~~\cite{madge2025mdc} for evaluation. 
The MDC\textendash R test split comprises 423 instances; each instance includes a scene image, a multi\textendash turn dialogue, and an associated target mention. As shown in Fig.~\ref {fig::case}, given an input of dialogue and image, the model predicts multiple bounding boxes that should refer to the ground truth blocks.
For training the model, we used three data sources corresponding to Sections \ref{sec:methodology:template}, \ref{sec:methodology:prompted} and \ref{sec:methodology:dialogue}: 
(i) template\textendash based data (Template), 
(ii) AI\textendash generated short REs (AI-Short), and 
(iii) AI\textendash generated dialogue data (AI-Dialogue). Additionally, we used a large-scale GREC dataset, gRefcoco~\cite{he2023grec}, and a dataset by prompting GPT-4.1 in a one-shot manner, GPT-4-One-Shot, as baselines.

\subsection{Bounding Boxes Reading}
\label{sec:methodology:annotation}
MDC-R ~\cite{madge2025mdc} has assigned a unique identifier to each block, composed of letters and Arabic numerals, e.g., A1. This ensures that all entity mentions within a dialogue refer to distinct combinations of blocks. 

Minecraft allows obtaining the pixel locations of the bounding boxes that surround the blocks. This way, we obtained all bounding boxes for each image. Reading the ID of each block from the image remains a challenge. To this end, we designed a method to read it. We recovered block letters via a render-and-compare scheme: for each block $X$ and candidate letter $L$, we re-render only $X$ with $L$ and compute a reconstruction error (MAE) on the block’s pixels.


With this procedure, for each block $k$ in the image, we obtained a ground truth triplet $(\mathbf{C}_k, \mathbf{B}_k, \mathbf{U}_k)$, where $\mathbf{C}_k$ denotes the block coordinate, $\mathbf{B}_k$ the bounding-box, and $\mathbf{U}_k$ a unique ID identifier. These annotations serve as filtering conditions for RE synthesis within section \ref{sec:methodology}.
\todo[disable]{V: Do the different synthetic datasets have different sizes? How big are they? It would be nice to have this info. \\ done, on Table 2}
\subsection{Implementation Details  and Metrics}
\label{sec:experiment:implementation}
\todo[disable]{(why use Qwen2-VL here)}
We employed Qwen2-VL-7B~\cite{wang2024qwen2}, approaching state-of-the-art over other LVLMs, and MDETR-longformer~\cite{madge2025mdc} as baseline models. Both Qwen2-VL (LoRA adapters) and MDETR-Longformer are adapted by supervised fine-tuning. For Qwen2-VL-7B, fine-tuning is performed using supervised fine-tuning (SFT) with LoRA, employing a learning rate of \(1\times10^{-4}\), a cosine learning-rate schedule, and three training epochs. LoRA adapters are applied to all layers, with training conducted in bfloat16 precision and a per-GPU batch size of 4. Fine-tuning and prompt details are available on GitHub.
For metrics, we computed the average F1 score and Prec@(F1=1, IoU$\geq 0.5$)~\cite{he2023grec} across all predicted samples and ground-truth samples. 

\subsection{Results}
\label{sec:experiment:result}
\todo[disable]{V: At this point, it would not be bad to remind the reader what the actual task the model has to do. Is it to identify the BB of the referred object in the image? It's not completely clear. \\ done in section 4.1}

\begin{table}[htb]
\centering
\resizebox{\columnwidth}{!}{
\begin{tabular}{l|c|c|c|r|r|r|r}
    \hline\hline
    \multirow{2}{*}{Model} & \multirow{2}{*}{Setting} & \multirow{2}{*}{Data} & \multirow{2}{*}{Size} & \multicolumn{2}{c|}{Full Dialogue} & \multicolumn{2}{c}{Mention Only}\\
    \cline{5-8}
    & & & & Mean F1 & Precision & Mean F1 & Precision\\
    \hline
    MDETR-Longformer & FT & gRefcoco~\cite{he2023grec}& 209k & 9.0 & 1.4 & - & -\\
    \hline
    \multirow{8}{*}{Qwen2-VL-7B} & ZS & - &- & 5.3 & 5.2 & 4.2 & 3.6\\
    & ICL & -& - & 0.5 & 0.5 & 0.5 & 0.5\\
     & FT & gRefcoco~\cite{he2023grec} & 209k & 19.1 & 13.5 &-&-\\
    & FT & GPT-4-One-Shot & 1K & 21.2 & 5.2 & 26.0 & 6.7\\ 
     & FT & Template & 19k & \textbf{42.8} & \textbf{24.5} & \textbf{45.2} & \textbf{27.8}\\
     & FT & Template & 1k & 18.6 & 13.5 & 25.7 & 19.1\\
     & FT & AI-Short & 1k & 22.8 & 15.7 & 28.9 & 15.2\\
     & FT & AI-Dialogue & 1k & 24.8 & 11.2 & 27.7 & 10.4\\
    \hline
\end{tabular}
\vspace{-10pt}
}
\caption{Results on MDC-R~\cite{madge2025mdc} benchmark, where ZS short for Zero Shot, FT short for Fine-Tune, and ICL short for In-Context-Learning. Metrics of mean F1 score (\%) and Precision@(F1=1 and IoU$\geq$0.5) in (\%) are measured for each model. For text inputs, Full Dialogue and Mention Only represent utilizing all dialogue with the mention and only the mention for models, respectively, which is illustrated in Fig.~\ref{fig::case}.}

\label{tab::comparison}
\end{table}
\todo[disable]{V: Seems strange that ICL would be worse than ZS? \\ Yes, the model can not utilize the demonstration properly. }
Table \ref{tab::comparison} presents a comparison of the model before and after fine-tuning. It's evident that the synthesis data significantly enhances the model's performance to varying degrees. 
\todo[disable]{V: This is the first mention of hierarchy? What is meant by this exactly?}
First, the table demonstrates that fine-tuning on the synthesized data gives superior results to training on out-of-domain data (e.g., gRefCOCO), despite using far fewer samples.
\todo[disable]{V: It's also not quite clear what the difference is between full dialogue and mention only. \\JX: Done on the Table 2 }
Second, by comparing Template data with other data, larger quantities of in-domain data produce greater improvements. However, at the same data size, AI-Short and AI-Dialogue yield superior results to Template data in Mean F1.
Furthermore, across all reported metrics, the Mention-only setting achieves stronger performance, indicating that the model struggles to reliably resolve antecedents for mentions when full dialogue context is provided. 

\begin{table}
\centering
\resizebox{\columnwidth}{!}{
\begin{tabular}{c|c|c|r|r|r|r}
    \hline\hline
    \multicolumn{3}{c|}{Combination of Data} & \multicolumn{2}{|c|}{Full Dialogue} & \multicolumn{2}{|c}{Mention Only}\\
    \hline
    Template & AI-Short & AI-Dialogue & Mean F1 & Precision & Mean F1 & Precision\\
    \hline
    \textcolor{green!60!black}{\(\checkmark\)} & \textcolor{green!60!black}{\(\checkmark\)} & & \textbf{44.7} & \textbf{25.7} & \textbf{45.6} & 27.6\\
    \textcolor{green!60!black}{\(\checkmark\)} & & \textcolor{green!60!black}{\(\checkmark\)}& 39.7 & 23.5 & 43.1 & 28.7\\
    & \textcolor{green!60!black}{\(\checkmark\)} & \textcolor{green!60!black}{\(\checkmark\)} & 26.0 & 13.1 & 39.8 & 24.0\\
    \textcolor{green!60!black}{\(\checkmark\)} & \textcolor{green!60!black}{\(\checkmark\)} & \textcolor{green!60!black}{\(\checkmark\)} & 39.7 & \textbf{25.7} & 43.8 & \textbf{29.7}\\
    \hline
\end{tabular}
}
\caption{The performance of Qwen2-VL fine-tuned on different synthetic data combinations.
}
\label{tab::combination}
\end{table}


To assess the combined impact of layered synthetic data, we compared models trained on different combinations of them (Table~\ref{tab::combination}). 
When combining Template data with others, the trained model exhibits a decline in performance. We attribute this gap to a distributional mismatch between dialogue text and the short-RE format used during prior training. While ingesting synthesized dialogues can transfer some useful knowledge, it introduces spurious correlations and learning biases relative to concise short-text supervision. Beyond the table, we compared combining gRefCOCO with other synthetic data, which resulted in small improvements over the results with AI-dialogue only, but no improvement overall with the rest of the data.

\begin{table}
\centering

\resizebox{0.4\columnwidth}{!}{
\begin{tabular}{c|c}
\hline\hline
Setting & Accuracy \\
\hline
Zero Shot & 72.6 \\
Fine-tuning & 77.6 \\
\hline
\end{tabular}

}
\caption{Generalization capability evaluation results of Qwen2-VL on the SK-VG easy-level split, comparing zero-shot and three-tier data fine-tuning settings.
}
\label{tab::generalization}
\vspace{-10pt}
\end{table}

To evaluate generalization potential, we tested Qwen2-VL on SK-VG~\cite{chen2023advancing}, which requires scene-knowledge reasoning, under zero-shot and fine-tuning of three-tier data (Table~\ref{tab::generalization}). Fine-tuning improves accuracy substantially on the easy-level split, indicating stronger grounding of appearance-based and relational expressions. Performance drops on the hard split, which requires purely scene-knowledge reasoning beyond our visual-dialogic-focus data.

\subsection{Quantitative and Qualitative Study}
\label{sec:experiment:quantitative}
To evaluate the generation quality of AI-Dialogue based on the method~\ref{sec:methodology:dialogue}, we evaluate it using automatic metrics. The dataset achieves CLIPScore = 0.261, perplexity (PPL) = 5.97, and Distinct-2 = 0.57 (global = 0.088), indicating moderate image--text alignment, good fluency, and high lexical diversity, and supporting meaningful grounding with diverse language generation. As AI-Dialogue contains metaphorical expressions, we additionally conduct a manual evaluation on 10\% of the data, obtaining CHAIR-i = 10.1\%, which suggests low hallucination. We note the presence of some semantically uninformative utterances and leave improved dialogue quality control to future work.

For the performance gap between Full Dialogue and Mention-only settings, in Table~\ref{tab::combination}, our analysis shows that Full Dialogue introduces context interference: irrelevant or ambiguous turns dilute visual grounding and cause error propagation (e.g., misinterpreting prior actions or cross-turn references). Mention-only inputs avoid these issues and align more directly with salient visual evidence, explaining their superior performance.

Further analysis on the degradation (Template + AI-short + AI-Dialogue V.S. Template + AI-short) in Table~\ref{tab::combination} reveals failure patterns when composing AI-Dialogue: (1) dialogue-history interference that overrides clear spatial cues, and (2) coreference amplification, where repeated mentions bias models toward duplicating bounding boxes for the same object. These effects quantitatively explain why the combined three-tier data underperform simpler compositions.

We presented qualitative results in Fig. \ref{fig::case}. As can be seen from the illustration, compared to the model without fine-tuning, the fine-tuned model considers the integrity of the target in inference and captures more details of the target.

\small\section{Conclusion}
\label{sec:conclusion}
This paper addresses GREC data scarcity stemming from the high cost of annotation. We achieved substantial performance gains via a three-tier data synthesis method, followed by model fine-tuning. The method is generalizable to other vision-language tasks facing limited supervision. Future work could adopt distribution-aware training to mitigate biases arising from heterogeneous REC and GREC data.

\small\section{Acknowledgements}
\label{sec:acknowledgements}
This research is funded by ARCIDUCA, EPSRC EP/W001632/1.



\bibliographystyle{IEEEtran}
\small\bibliography{main}

\end{document}